\documentclass[final,3p]{elsarticle}

\usepackage{amsmath}
\usepackage{amssymb}
\usepackage{booktabs}
\usepackage[colorlinks=true,breaklinks=true,pdftex]{hyperref}
\biboptions{authoryear}

\begin{document}

\begin{frontmatter}

\title{Text-Image Conditioned Diffusion for Consistent Text-to-3D Generation}

\author[first]{Yuze He}
\ead{hyz22@mails.tsinghua.edu.cn}
\author[first]{Yushi Bai}
\author[first]{Matthieu Lin}
\author[first]{Jenny Sheng}
\author[first]{Yubin Hu}
\author[first]{\\Qi Wang}
\author[bjtu]{Yu-Hui Wen}
\author[first]{Yong-Jin Liu\corref{cor}}
\ead{liuyongjin@tsinghua.edu.cn}
\cortext[cor]{Corresponding author}
\address[first]{Department of Computer Science and Technology, Tsinghua University}
\address[bjtu]{School of Computer and Information Technology, Beijing Jiaotong University}

\begin{abstract} 
By lifting the pre-trained 2D diffusion models into Neural Radiance Fields (NeRFs), text-to-3D generation methods have made great progress. 
Many state-of-the-art approaches usually apply score distillation sampling (SDS) to optimize the NeRF representations, which supervises the NeRF optimization with pre-trained text-conditioned 2D diffusion models such as Imagen. 
However, the supervision signal provided by such pre-trained diffusion models only depends on text prompts and does not constrain the multi-view consistency. 
To inject the cross-view consistency into diffusion priors, some recent works finetune the 2D diffusion model with multi-view data, but still lack fine-grained view coherence. 
To tackle this challenge, we incorporate multi-view image conditions into the supervision signal of NeRF optimization,
which explicitly enforces {\bf fine-grained view consistency}. 
With such stronger supervision, our proposed text-to-3D method effectively mitigates the generation of floaters (due to excessive densities) and completely empty spaces (due to insufficient densities). 
Our quantitative evaluations on the T$^3$Bench dataset demonstrate that our method achieves state-of-the-art performance over existing text-to-3D methods. We will make the code publicly available.
\end{abstract}


\end{frontmatter}

% Comment out for final accepted paper submission
% \linenumbers

%%%%%%%%%%%%%%%%%%%%%%%%%%%%%%%%%%%%%%%%%%%%%%%%%%%%%%%%%%%%%%%%%%%%%

\section{Introduction}

The advent of text-to-3D technology marks a revolutionary stride in digital asset generation, profoundly impacting various fields, such as gaming and other 3D applications involving a creative design process. 
In the gaming industry, this technology heralds a new era of creativity and efficiency, allowing designers to rapidly transform textual descriptions into intricate 3D models, thereby reducing development time and fostering more dynamic and immersive gaming experiences. 
Recently, researchers are increasingly concentrating on improving the geometric priors of large-scale diffusion models for generative text-to-3D tasks.

\begin{figure}[htb]
\centering
\includegraphics[width=0.9\linewidth]{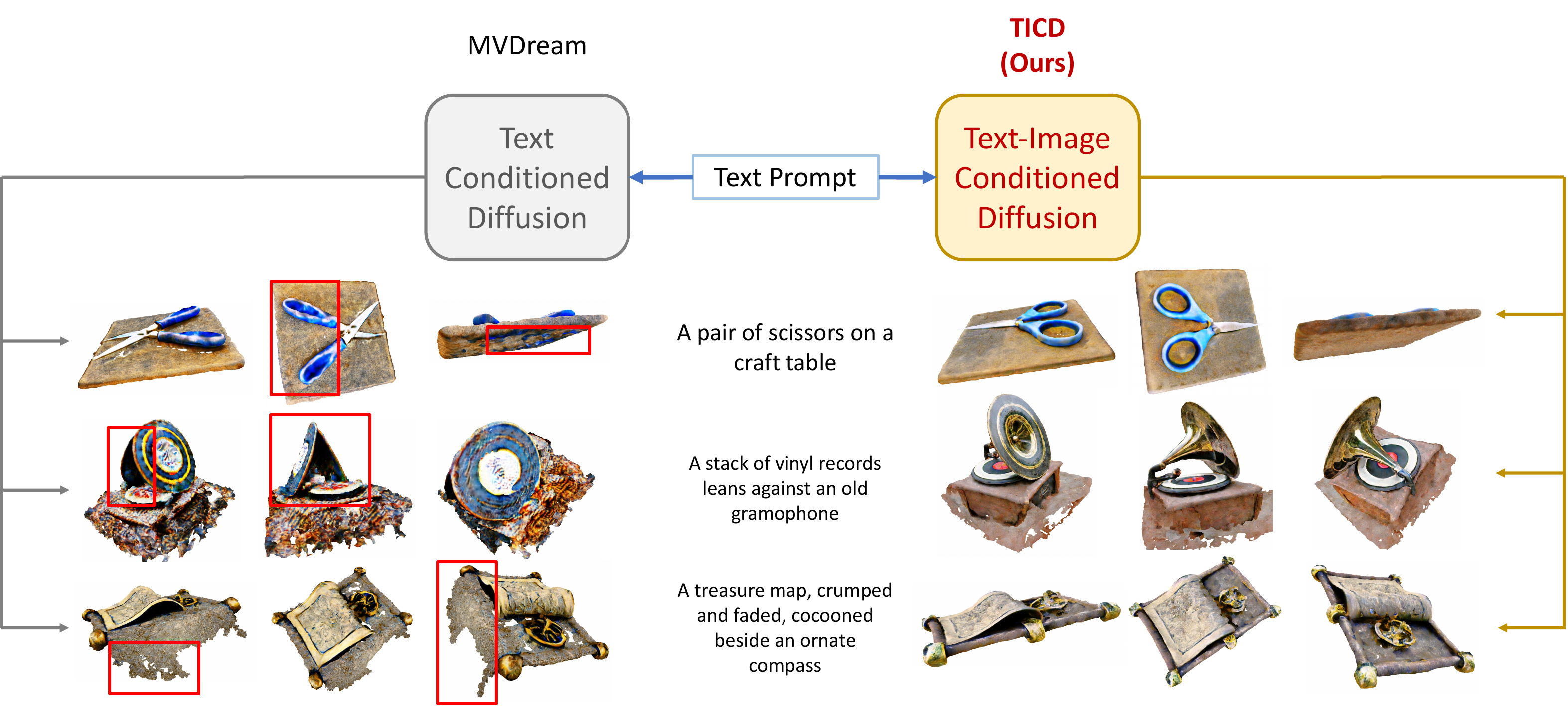}
\caption{A few qualitative comparison examples of our method (TICD) with a state-of-the-art method MVDream (\cite{mvdream}). Our method uses text-image conditioned diffusion, which incorporates multi-view image conditions into the supervision signal of NeRF optimization, helping generate 3D contents with fine-grained consistency and clear geometry. Our method consistently performs better than MVDream and other state of the arts on T$^3$Bench; see Tab.~\ref{tabl:t3} in Sec.\ref{sec:experiment} for quantitative comparisons.}
\label{fig:teaser}
\end{figure}

Prominent methods in text-to-3D generation demonstrate significant strengths and potential in two steps: they first implement the generative model as a differentiable mapping from parameters to images (\cite{dreamfusion, mvdream, prolificdreamer, magic3d});
and then a neural inverse renderer, NeRF (\cite{nerf}), is trained to render images that belong to the distribution of the pre-trained diffusion model. 
In other words, this class of methods distill the score function of a pre-trained diffusion model, namely score distillation (\cite{dreamfusion}).
However, the effective training of a neural inverse renderer necessitates consistent multi-view images, a requirement current diffusion models struggle to meet due to their inherent stochasticity (\cite{dreamfusion, magic3d, textmesh}). Consequently, the resulting objects often suffer from the Janus problem (\cite{dreamfusion}), which refers to as regenerating multi-copy (and thus wrong) content described by the text prompt like multi-faces or drifted content. 

Additionally, since the generative process is inherently a 3D reconstruction from 2D observations, the problem is highly ill-posed.
Without enough consistency between views to constrain the reconstruction, NeRF tends to predict excessive or insufficient densities.
Excessive densities correspond to an object being visually occluded, and insufficient densities result in failure to generate transparent objects.
Overall, achieving both appropriate densities and high color consistency is extremely challenging for the stability of the 3D generation task.

Recently, ProlificDreamer (\cite{prolificdreamer}) takes into account the stochastic process during optimization, yielding high-quality 3D assets. 
However, the optimization process is prohibitively expensive, making it impractical for real-time applications.
Other state-of-the-art works like MVDream (\cite{mvdream}) finetune the diffusion model to yield multi-view images by generating a set of orthogonal views with respect to the text prompt. 
Nevertheless, MVDream does not always produce consistent multi-view images, as its fine-tuning process involves generating views with minimal overlap with the text. 
Consequently, the optimization process does not provide enough constraints, and the resulting NeRF tends to predict excessive or insufficient densities on certain text prompts. 
To address this challenge, in this paper, we consider whether it can give sufficient constraints to control the diffusion models to generate high-quality objects.

Drawing on the above considerations, we propose to add an additional image-conditioned diffusion model to constrain the 3D reconstruction during score distillation, as shown in Figure~\ref{fig:teaser}.
The additional guidance enforces fine-grained view consistency in the rendered images.
We start by sampling a set of orthogonal views, which we refer to as reference views.
Then, we randomly sample a reference view and use it as a condition for the image-guided diffusion model to produce a novel view. 
Furthermore, we apply the score distillation as evaluated by two different diffusion models.
On the one hand, a text-conditioned multi-view diffusion model gives an updated direction following multi-view consistency with respect to text (which indicates accuracy), constraining the coarse consistency of the 3D reconstruction;
On the other hand, the image-conditioned novel view diffusion model gives an updated direction following consistency between views, which can enforce the fine-grained consistency between views instead of the text (which indicates quality). 
Finally, the two complementary signals collaborate to ensure the accuracy and quality of the 3D target generation.
Our experimental results on T$^3$Bench show that our method outperforms existing methods and yields state-of-the-art results. 

We organize this paper as follows. First, we present related work with score distillation in Sec. \ref{sec:relatedwork}. 
Then, we describe our method in Sec. \ref{sec:method}, which leverages an image-conditioned diffusion model to explicitly enforce multi-view consistencies.
Experimental results are shown in Sec. \ref{sec:experiment}. 
Finally, we conclude our work in Sec. \ref{sec:conclusion}.

\section{Related Work}
\label{sec:relatedwork}

\subsection{Text-to-3D}
With the rise of large pre-trained 2D models (\cite{clip, imagen, stablediffusion, if}), text-to-3D generation has experienced a remarkable leap in progress. Existing text-to-3D methods can be grouped into two main approaches. The first group of works (\cite{shap-e, point-e, 3dgen}) adopts only captioned 3D data for the text-to-3d generation task. However, due to the fact that captioned 3D data is limited in scale and diversity, these methods are very dependent on access to large-scale 3D data and can only generate objects from a limited set of categories.

A second well-used paradigm is to tackle 2D supervision to supervise a 3D representation model such as NeRF (\cite{nerf}). Early works use CLIP (\cite{clip}) to optimize NeRF by projecting the text and images of the 3D generation into a shared latent space and aligning them with the CLIP loss (\cite{dreamfields, clip-mesh, dream3d, pureclip, text2mesh, taps3d}). Because CLIP is not a generative model, these methods tend to produce generations that lack geometric fidelity. Improving upon previous works, \citet{dreamfusion, sjc} are the pioneer works that leverage 2D pre-trained diffusion models for text-to-3D generation. They propose score distillation sampling (SDS) that supervises 3D representation models, such as NeRF (\cite{nerf}) or DMTet (\cite{dmtet}), by guiding their corresponding rendered images towards areas of high probability density that are contingent on the text. Expanding upon the aforementioned research based on SDS, existing works explore different supervision methods (\cite{prolificdreamer, csd}) and various 3D representations such as Gaussian Splatting (\cite{cg3d, dreamgaussian, gaussiandreamer, gsgen}).

Building upon these seminal works, a large collection of 2D-lifting methods (\cite{dreamtime, prolificdreamer, csd, dittonerf, textmesh}) focus on improving fidelity and ameliorating existing issues, such as over-saturation (generations tend to appear cartoon-like). In addition, prior works also tackle the poor view-consistency issue and the Janus problem (\cite{mvdream, efficientdreamer, hifa, boosting3d, geodream, sweetdreamer, 3dfuse, debiasing}). Moreover, efficiency-centric methods emphasize reducing the generation time through parallel sampling (\cite{dreampropeller}), distilling a multi-view 2D diffusion model (\cite{et3d}), and splitting the task into sparse view generation and reconstruction (\cite{instant3d}). Other methods (\cite{magic3d, fantasia3d}) focus on generating high-resolution 3D contents using a coarse-to-fine approach and exploring efficient 3D representations, as well as producing generations with fine details (\cite{richdreamer}). Several other works improve generation diversity (\cite{dreamtime, prolificdreamer, dittonerf, direct25}). 

\subsection{Image-to-3D}
NeRF (\cite{nerf}), with its ability to take a collection of images from different viewpoints and reconstruct the underlying 3D scene, has shown tremendous potential in progressing the image-to-3D research direction. However, NeRF is greatly constrained by the large demand for dense viewpoint data with significant viewpoint overlaps. Subsequent works focus on relaxing this constraint under the single-image-to-3D paradigm, such as performing generation through progressively deforming the mesh (\cite{pixel2mesh}), leveraging NeRF with learned scene prior (\cite{pixelnerf}), adding semantic/geometric regularization (\cite{sinnerf}), or leveraging differentiable surface modeling and differentiable rendering (\cite{get3d}). However, these methods are limited by the lack of large-scale 3D data as well as suffer from numerous issues such as inconsistency and long generation times.

Later works leverage 2D models as priors in the 3D generation process (\cite{nerdi}) and take a coarse-to-fine approach (\cite{makeit3d, dittonerf, magic123}. Subsequent methods likewise adopt the 2D diffusion prior to the novel view synthesis task (\cite{3dim, genvs, sparsenerf, zero123, visiontransformer}). Expanding upon established foundations, the current image-to-3D task features several orthogonal research focuses, include enabling 360-degree reconstruction of objects (\cite{realfusion, neurallift360}) and scenes (\cite{zeronvs}), improving generation efficiency (\cite{12345, 12345++}), and ameliorating 3D inconsistencies (\cite{consistent1to3, consistent123caseaware, consistent123, wonder3d, syncdreamer}).

\subsection{Multi-view Diffusion} 
One major approach to ameliorate the multi-view consistency issue in 2D-lifting methods is through leveraging multi-view diffusion models, which requires modeling the joint distribution of an object's information from multiple views. The multi-view information can take the form of latent intermediary features (\cite{mvdiffusion}), epipolar lines (\cite{ poseguideddiff}), or whole images (\cite{mvdream}). Information-sharing is achieved with a variety of techniques, such as placing different viewpoint images into a tiling layout (\cite{zero123++, textmesh, instant3d}), denoising several image views with different noises and then sharing information across images with an attention mechanism during each denoising step (\cite{mvdream, syncdreamer,mvdiffusion, dmv3d}), or adopting a coarse-to-fine approach by incorporating an orthogonal-view diffusion prior (\cite{efficientdreamer}).

%%%%%%%%%%%%%%%%%%%%%%%%%%%%%%%%%%%%%%%%%%%%%%%%%%%%%%%%%%%%%%%%%%%%%
\section{Our Text-Image Conditioned Diffusion (TICD) Method}
\label{sec:method}
In Sec. \ref{sec:preliminaries}, we provide some preliminaries that are necessary to our method.
In Sec. \ref{sec:ourmethod}, we present a novel score distillation that leverages text-conditioned and image-conditioned score functions.
We propose to constrain the optimization process to generate multi-view images that are consistent both with the text and across different views. The detailed implementation is presented in Sec. \ref{subsec:details}.

\subsection{Preliminaries}
\label{sec:preliminaries}

\textbf{NeRF} (\cite{nerf}) is a neural inverse rendering approach based on volume rendering.
In formulae, for each point in space and viewing direction unit vector in $\mathbb{R}^{3}$, NeRF is a differentiable volumetric renderer parameterized by a neural network $\phi$ that returns the density $\sigma$ and RGB color $\mathbf{c}$.
In particular, NeRF renders each pixel via volume ray casting. 
Given a ray $\mathbf{r}(t)= \mathbf{o} + t\mathbf{d}$ with ray origin $\mathbf{o}$ the camera center, it approximates the following integral via numerical quadrature:
\begin{equation}
    C(\mathbf{r}) = \int_{t_n}^{t_f}T(t)\sigma_{\phi}(\mathbf{r}(t))\mathbf{c}_{\phi}(\mathbf{r}(t),\mathbf{d})dt,
    \label{eq:render}
\end{equation}
where $T(t)$ is the transmittance function that predicts the probability that the ray travels from $t_n$ and terminates at $t_f$.
NeRF is trained in an end-to-end manner on a set of posed images to minimize the reconstruction loss between the rendered color and the ground-truth color of the posed images.
For simplicity, denote $x=g(\phi, c)$ a rendered view at some desired camera pose $c$, which is obtained by applying Eq. \ref{eq:render} on every pixel.

\textbf{DDPM} (\cite{ddpm}) is described as two Markov processes yielding latent variables $\{z_t\}_{t=1}^T$ from a data distribution $x\sim p_{\text{data}}$. Specifically, it defines a forward process $q(z_t\mid z_{t-1})$ and a reverse process $p(z_{t-1}\mid z_t)$, both being Gaussians. While $q$ is hand-crafted, we learn $p$ using a deep neural network (DNN) with parameters $\theta$, which we denote $p_{\theta}$. In formulae, given $x$ the diffusion process $q$ generates latent variables $\{z_t\}_{t=1}^T$ with decreasing signal-to-noise ratio such that: 

\begin{equation}
    q(z_t \mid x) = \mathcal{N}(z_t, \sqrt{\bar{\alpha}_t}x,(1-\bar{\alpha}_t)I),
    \label{eq:q}
\end{equation}

where $\alpha=1-\beta_t$, $\bar{\alpha}_t=\prod_{i=0}^t\alpha_i$, and $\beta_t$ is a hyper-parameter that controls the noise level at $t$.
Since $q(z_t)$ converges towards an isotropic Gaussian distribution as $t\rightarrow \infty$, we define the prior of the reverse process $p_{\theta}$ as an isotropic Gaussian distribution, $p_{\theta}(z_T) = \mathcal{N}(0, I)$. Given a noising schedule $\sigma^2_t \propto \beta_t$, the reverse process runs from $t=T$ to $t=1$ and at each step we sample from $\mathcal{N}(\mu_{\theta}(z_t, t), \sigma_t^2I)$ where: 

\begin{equation}
\mu_{\theta}(z_t, t) = \frac{\sqrt{\alpha_{t-1}}\beta_t }{1-\alpha_t}x_{\theta}(z_t,t) + \frac{\sqrt{1-\beta_t}(1-\alpha_{t-1})}{1-\alpha_t} z_t.
\end{equation}

In particular, the DNN $x_{\theta}(z_t, t)$ predicts the signal $x$ such that it minimizes the following loss $J(\theta)$ defined as

\begin{equation}
    J(\theta) = \mathbb{E}_{x,t}\left[w(t) \lVert x - x_{\theta}(z_t, t)\rVert^2_2\right]
    \label{eq:ddpmobjective},
\end{equation}
where $w(t)$ is a positive weighting function that depends on $t$ and $z_t$ is obtained using Eq. \ref{eq:q}.
Additionally, one can use classifier-free guidance to decrease the diversity of the samples while increasing the quality of each individual sample w.r.t to some condition $y$ (e.g., text).
Notably, the neural network's output is modified as follows
\begin{equation}
    \tilde{x}_{\theta}(z_{t}, c)
=
\gamma
x_{\theta}(z_{t}, y)
+
(1-\gamma)
x_{\theta}(z_{t}),
\label{eq:cfg}
\end{equation}
where $\gamma$ is a scalar coefficient that balances the aforementioned trade-off.

\begin{figure}[t]
\centering
\includegraphics[width=0.9\linewidth]{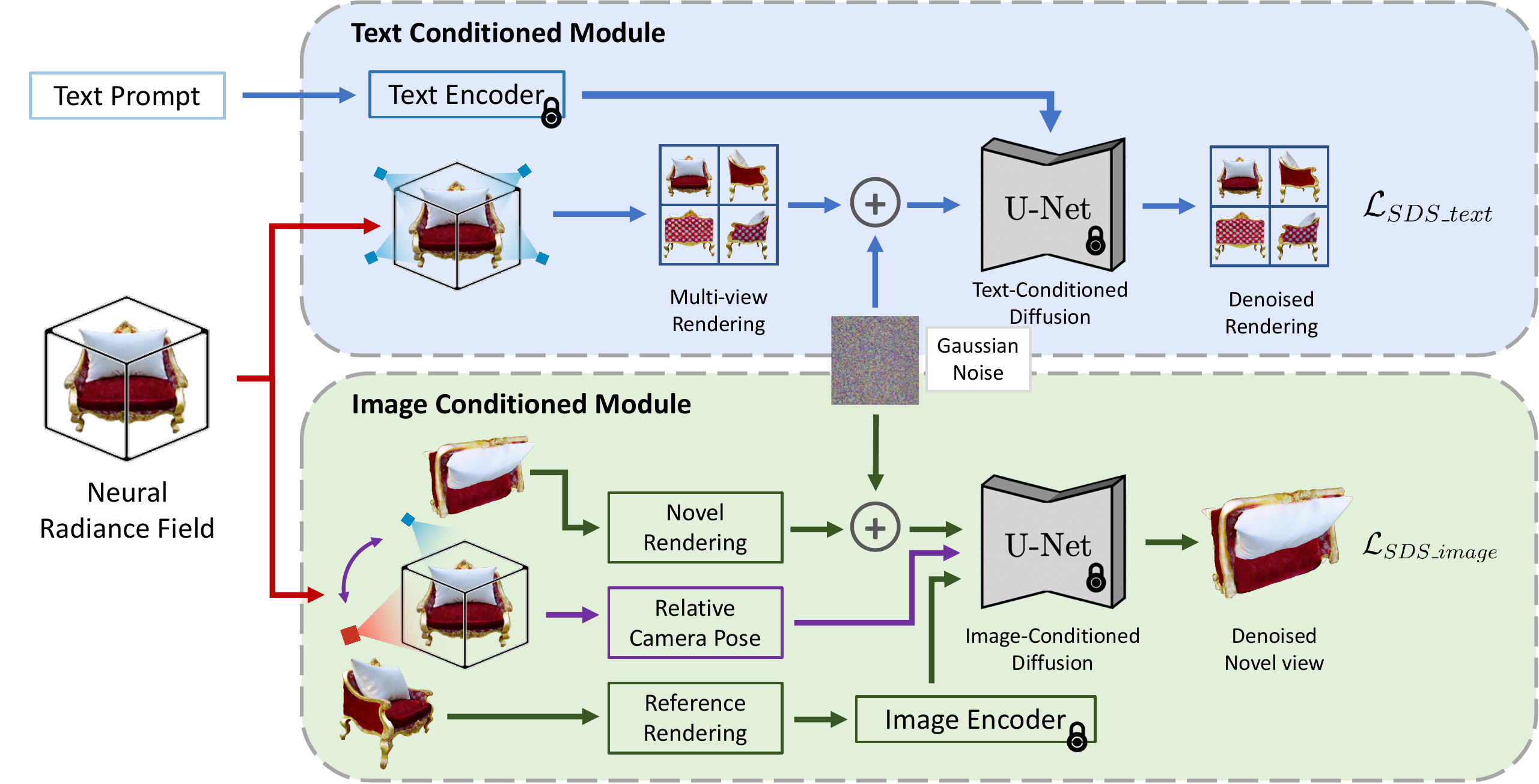}
\caption{The pipeline of our proposed method. Taking a text prompt as input, we optimize the neural radiance field by using both text and image-conditioned modules as input. The image-conditioned module applies extra progressive consistency constraints that help generate 3D contents with consistent and accurate geometry. }
\label{fig:pipe}
\end{figure}

\subsection{Image and Text View Consistency}
\label{sec:ourmethod}

Neural inverse rendering techniques such as NeRF optimize a 3D scene representation based on image observations. 
Thus, one can use a text-to-image generative model for optimizing a 3D scene representation.
Prior work refers to this as score distillation sampling (\cite{dreamfusion}), for the sampling process of this generative model is done by distilling the learned score function of a 2D diffusion model into NeRF.
Leveraging text-to-image generative models has the following advantages. 
First, compared to 3D data, there is abundant 2D data. 
Second, it allows leveraging other abilities in 2D generative models, such as editing or controllability (\cite{mvcontrol, instructnerf2nerf, control3d}).
Finally, it can also be applied in a similar manner to video generation (\cite{textto4d}).  
In practice, we generate multi-view images using a text-to-image diffusion model and minimize the score distillation loss on these images by optimizing NeRF.

Score Distillation (\cite{dreamfusion}) encourages NeRF to render an image $x$ such that it belongs to the distribution of plausible images as evaluated by a diffusion model $\hat{x}_{\theta}(\cdot)$. In practice, it consists of optimizing $\phi$ to minimize the following loss function:
\begin{equation}
    J(\phi)
    =
    % \mathbb{E}_{t, y, c}\left[
    % \lVert \hat{x}_{\theta}(z_t; y, c, t) - x \rVert_2^2
    \mathbb{E}_{x, t, y}\left[
    \lVert x - \hat{x}_{\theta}(z_t; y, t) \rVert_2^2
    \right],
    \label{eq:sds}
\end{equation}
where $z_t$ is sampled from the forward process and $y$ is some condition (e.g., text). In practice, the gradient of the diffusion model is detached, yielding the following gradient for NeRF's parameters $\phi$:
\begin{equation}
    \nabla_{\phi}J(\phi)
    =
    \mathbb{E}_{x, t, y}\left[
     \left(x - \hat{x}_{\theta}(z_t; y, t) \right)
     \frac{\partial x}{\partial \phi}
    \right]
    \label{eq:grad}.
\end{equation}
Eq. \ref{eq:render} shows that the optimization process in NeRF is applied in a pixel-wise manner.
Thus, pixel-wise view consistency is the signal for learning the density and color.
Additionally, the gradient signal in Eq. \ref{eq:grad} must maintain multi-view consistency to match the condition $y$.
In other words, accurate text-to-3D asset generation requires view consistency w.r.t $y$ and between views at the pixel level.

Our score distillation uses two different scores for supervision: a text-conditioned multi-view diffusion model and an image-conditioned novel-view diffusion model.
These two learning signals are complementary; one maintains multi-view consistency with text, while the other maintains consistency between views. 
In this paper, we denote a view as $x$ and a set of views (i.e., multi-view) as $\mathbf{x}$ where we index each view as $\mathbf{x}^i$. 

\textbf{Text-conditioned generation.} Denote with $i$ the index of views generated from the text condition. 
Following prior work (\cite{mvdream}), we start by sampling a set of camera poses $\mathbf{c}$ and render these views $\mathbf{x} = g(\phi, \mathbf{c})$, which we call reference views.
In particular, the views $\mathbf{x}$ are chosen such that they are orthogonal to each other. 
For each view, we sample a timestep $t$ and compute the forward process of the diffusion process $\mathbf{z}_t^i$.
Given the text $y$ and the set of noised views $\mathbf{z}_t$ rendered from NeRF, the text-conditioned diffusion model $\hat{x}_{\theta_1}(\mathbf{z}_t; y, \mathbf{c}, t)$ computes score function w.r.t to $\mathbf{z}_t$, yielding an update direction towards higher density regions.
Additionally, we modify the score function to include classifier-free guidance as in Eq. \ref{eq:cfg} and compute score distillation to obtain gradient update as in Eq. \ref{eq:grad}.

\textbf{Image-conditioned generation.} Prior work utilized image-conditioned diffusion to generate 3D content from a single 2D reference image, while our approach {\it differs in how the model is leveraged}. Instead of providing the reference as an additional input to the model, we innovatively employ it as extra supervision to guide different views and ensure fine-grained multi-view consistency.
Denote with $j$ the index of novel views generated from images.
We render extra views $\mathbf{x}^j$ at camera extrinsic $\mathbf{c}^{j}$.
Denote the relative camera extrinsic $\mathbf{c}^{(j\rightarrow i)}$ from camera position $i$ to $j$. 
In formulae, the image-conditioned diffusion model takes rendered image $\mathbf{x}^j$ along with the relative camera extrinsic $\mathbf{c}^{(j\rightarrow i)}$ as conditioning.
In a similar manner, we sample $t$ from the uniform distribution. 
The model was trained to compute the score function for novel views $\mathbf{z}^i_t$, denoted as $\hat{x}_{\theta_2}(\mathbf{z}_t^i;\mathbf{x}^j, \mathbf{c}^{(j\rightarrow i)}, t)$.
Following \cite{zero123}, we compute the score distillation at $\mathbf{z}^i_t$, which was previously rendered. 
The resulting score distillation encourages NeRF to render consistent views $\mathbf{x}^j$, such that the pre-trained diffusion models can predict accurate novel views $\mathbf{x}^i$.

\textbf{Text and image score distillation.} We modify Eq. \ref{eq:sds}, such that the gradient signal becomes:
\begin{equation}
    \nabla_{\phi}J(\phi)=
    \mathbb{E}_{x, t, y, }\left[
    \left\{
    \lambda_t
    \left(\underbrace{\hat{x}_{\theta_1}(\mathbf{z}_t; y, \mathbf{c}, t)}_{\text{text diffusion model}} - \mathbf{x}\right)
    +
    \lambda_i
    \left(\underbrace{\hat{x}_{\theta_2}(\mathbf{z}_t^i;\mathbf{x}^j, \mathbf{c}^{(j\rightarrow i)}, t)}_{\text{image diffusion model}} - \mathbf{x}^j\right)
    \right\}
     \frac{\partial x}{\partial \phi}
    \right]
    \label{eq:loss},
\end{equation}
where $\lambda_t$ and $\lambda_i$ are scale factors of the text and image diffusion model, respectively. The score distillation process adds a score toward view consistency and another score toward text multi-view consistency.
This optimization process provides NeRF with enough constraints to predict accurate densities and colors as shown in Figure \ref{fig:abl}.

\subsection{Implementation Details}
\label{subsec:details}

\textbf{Models and Representations.} We utilize MVDream's (\cite{mvdream}) pre-trained model as our multi-view diffusion model, and Zero123-xl, provided by Zero-1-to-3 (\cite{zero123}), as our novel view image-conditioned diffusion model. For 3D representation, we implement ThreeStudio's (\cite{threestudio2023}) implicit volume approach, consisting of a multi-resolution hash grid and an MLP network for predicting voxel density and RGB values.

\textbf{View Selection.} For each camera view to render, we first randomly select cameras with a field-of-view (fov) between [15, 60] and an elevation between [0, 30] for the multi-view diffusion model. The camera distance is set as the object size (0.5) multiplied by the NDC focal length and a random scaling factor ranging from [0.8, 1.0]. We then randomly select views from the above set as reference views for the novel view diffusion model. For each reference view, we choose an additional random camera with the same fov and an elevation between [-30, 80] before applying the novel view image-conditioned diffusion model. The batch size starts at 8 and 12 for the multi-view and novel view models, respectively, and then decreases to 4 and 4 after 5,000 iterations.

\textbf{Optimization.} The 3D model is optimized for 10,000 steps using an AdamW (\cite{kingma2014adam}) optimizer. The learning rate for the hash-grid and MLP components is set to 0.01 and 0.001, respectively. Score distillation sampling is applied, with the maximum and minimum time steps decreasing from 0.98 to 0.5 and 0.02 over the first 8,000 steps, respectively. Both the loss scale factors $\lambda_t$ and $\lambda_i$ are set to 1.0. The rendering resolution begins at 64$\times$64 and is increased to 256$\times$256 after 5,000 steps. Guidance scales of 50.0 and 3.0 are used for the multi-view and novel-view models, respectively.

\section{Experiment}
\label{sec:experiment}

\subsection{Dataset and Settings}
We evaluate our method on the T$^3$Bench (\cite{he2023t}), a comprehensive text-to-3D benchmark containing diverse text prompts across three categories: {\it Single Object}, {\it Single Object with Surroundings}, and {\it Multiple Objects}, with 100 distinct prompts in each category.
We also leverage the two automatic metrics proposed in T$^3$Bench (\cite{he2023t}) for evaluating the quality and text alignment of generated 3D scenes. The quality metric captures multi-view scene images and utilizes text-image scoring models and a regional convolution mechanism to measure overall quality and detect view inconsistency issues. The alignment metric employs multi-view scene captioning with BLIP (\cite{li2022blip}) and aggregates the captions with a large language model, which measures how well the caption covers information in the original text prompt. 

Apart from the analysis of 3D content, we also conduct a text-2D consistency analysis by applying the CLIP (\cite{clip}) cosine similarity between the original text prompt and the image captured at a fixed position and orientation. In practice, we first normalize the 3D mesh into a cube with a $[-1, 1]$ range, then capture an image at the front of the 3D mesh, with a distance of $2.2$ and a focal of $3.0$.

For extracting the mesh model, we adopt the Marching Cubes (\cite{lorensen1998marching}) algorithm for our method and the original mesh extracting algorithm in T$^3$Bench (\cite{he2023t}) for other methods. We also follow the settings of T$^3$Bench to apply mesh geometry simplification to a maximum of 40,000 faces before texture extraction.

\begin{table}[t]
\centering
% \resizebox{1.0\linewidth}{!}{
\begin{tabular}{lccc}
\toprule
 & Single Obj. & Single Obj. w/ Surr. & Multiple Obj. \\
\midrule
DreamFusion & 24.4 & 24.6 & 16.1 \\
Magic3D     & 37.0 & 35.4 & 25.7 \\
LatentNeRF  & 33.1 & 30.6 & 20.6 \\
Fantasia3D  & 26.4 & 27.0 & 18.5 \\
SJC         & 24.7 & 19.8 & 11.7 \\
ProlificDreamer & 49.4 & 44.8 & 35.8 \\
MVDream     & 47.8 & 42.4 & 33.8 \\
TICD  &   \textbf{50.0}    & \textbf{45.6}  &	\textbf{36.0} \\
\bottomrule
\end{tabular}
% }
\caption{Average scores (\%) on T$^3$Bench (\cite{he2023t}).}
\label{tabl:t3}
\end{table}

\begin{table}[h]
\centering
% \resizebox{1.0\linewidth}{!}{
\begin{tabular}{lccc}
\toprule
 & Single Obj. & Single Obj. w/ Surr. & Multiple Obj. \\
\midrule
DreamFusion & 0.255	& 0.253	& 0.240 \\
Magic3D     & 0.264	& 0.268	& 0.253 \\
LatentNeRF  & 0.271	& 0.267	& 0.247 \\
Fantasia3D  & 0.250	& 0.248	& 0.240 \\
SJC         & 0.245	& 0.232	& 0.233 \\
ProlificDreamer & 0.283	& 0.261	& 0.251  \\
MVDream     & 0.281	& 0.277	& 0.265 \\
TICD        & \textbf{0.285}	& \textbf{0.279}	& \textbf{0.268} \\
\bottomrule
\end{tabular}
% }
\caption{CLIP cosine similarities between the front view image and the original text prompt.}
\label{tabl:clip}
\end{table}

\subsection{Comparisons with State-of-the-Art}
We compared our method with the seven advanced text-to-3D methods, including Dreamfusion (\cite{dreamfusion}), Magic3D (\cite{magic3d}), LatentNeRF (\cite{metzer2023latent}), Fantasia3D (\cite{fantasia3d}), SJC (\cite{sjc}), ProlificDreamer (\cite{prolificdreamer}), and MVDream (\cite{mvdream}). For all the methods, we adopt the original implementation of ThreeStudio (\cite{threestudio}) to generate 3D contents, and the original codebase of T$^3$Bench to extract 3D meshes and evaluate metrics. 

\begin{table}[t]
\centering
% \resizebox{1.0\linewidth}{!}{
\begin{tabular}{lccc}
\toprule
 & Single Obj. & Single Obj. w/ Surr. & Multiple Obj. \\
\midrule
without image condition & 48.5    &   43.2    &   34.0 \\
with image condition  &   \textbf{50.0}    & \textbf{45.6}  &	\textbf{36.0} \\
\bottomrule
\end{tabular}
% }
\caption{Results of ablation study on image-conditioned generation.}
\label{tabl:abl}
\end{table}

\begin{figure}[htbp]
\centering
\includegraphics[width=0.9\linewidth]{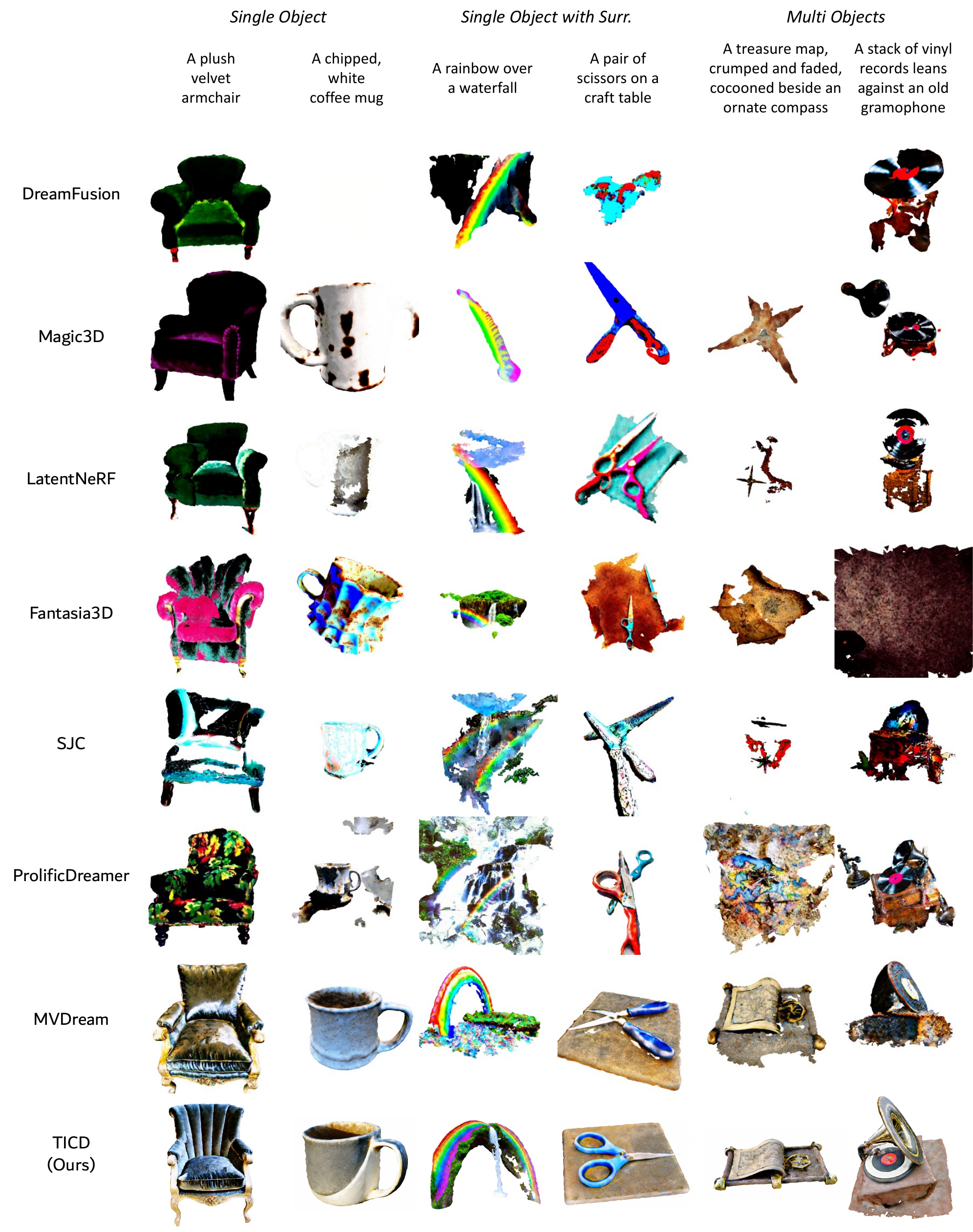}
\caption{Qualitative comparisons between our method and other advanced methods. The 3D content generated by our method shows better quality, clearer geometry, and higher consistency.}
\label{fig:main}
\end{figure}

Our method achieves the best performance among all the text-to-3D methods on the average T$^3$Bench score across all the prompt sets, as shown in Table~\ref{tabl:t3}. On the one hand, compared with the performance of a single object, our precision of reconstruction is significantly higher than other methods, proving our superiority. On the other hand, for the more difficult multiple objects task, our method shows a strong result, which is attributed to the co-supervision of text and image features.

As illustrated in Figure~\ref{fig:main}, the 3D-generated content from most existing methods lacks texture details and clear geometry, making it difficult to faithfully reproduce all information from textual prompts. DreamFusion produces renderings lacking in fine texture details and suffers from a high failure rate. Magic3D and LatentNeRF generate improved texture details but still demonstrate poor geometric quality. SJC exhibits a tendency to output less compact geometry, an attribute unfavorable to the final shape of the 3D content. By contrast, Fantasia3D typically outputs compact and well-defined geometry alongside richer textures. However, its performance declines when processing complex textual prompts, often yielding completely erroneous geometry.
While ProlificDreamer utilizes LoRA (\cite{hu2021lora}) to finetune the diffusion model during optimization and variational score distillation, which benefits the generation of rich details, the geometry quality is often poor with excessively incorrect shapes. MVDream performs much better geometrically, but the overall generation quality and alignment to the original text prompt need further improvement. Our proposed method demonstrates superior performance across all aspects; it can generate detailed 3D content with accurate, high-quality geometry that effectively reflects the textual description, even on challenging prompt sets.

To further validate the effectiveness, we also construct the CLIP cosine similarities comparison test between the front view image and the original text prompt, as shown in Table~\ref{tabl:clip}. The results also support that our method retains better consistency with the original text prompt compared to other methods.

\subsection{Ablation Study}
To demonstrate the effectiveness of the image-conditioned diffusion module, we first conducted quantitative experiments, adjusting only the inclusion of the module while keeping all other experimental conditions constant. As shown in Table~\ref{tabl:abl}, the addition of the image-conditioned module effectively improved response quality and alignment with the original text prompt.

\begin{figure}[t]
\centering
\includegraphics[width=0.8\linewidth]{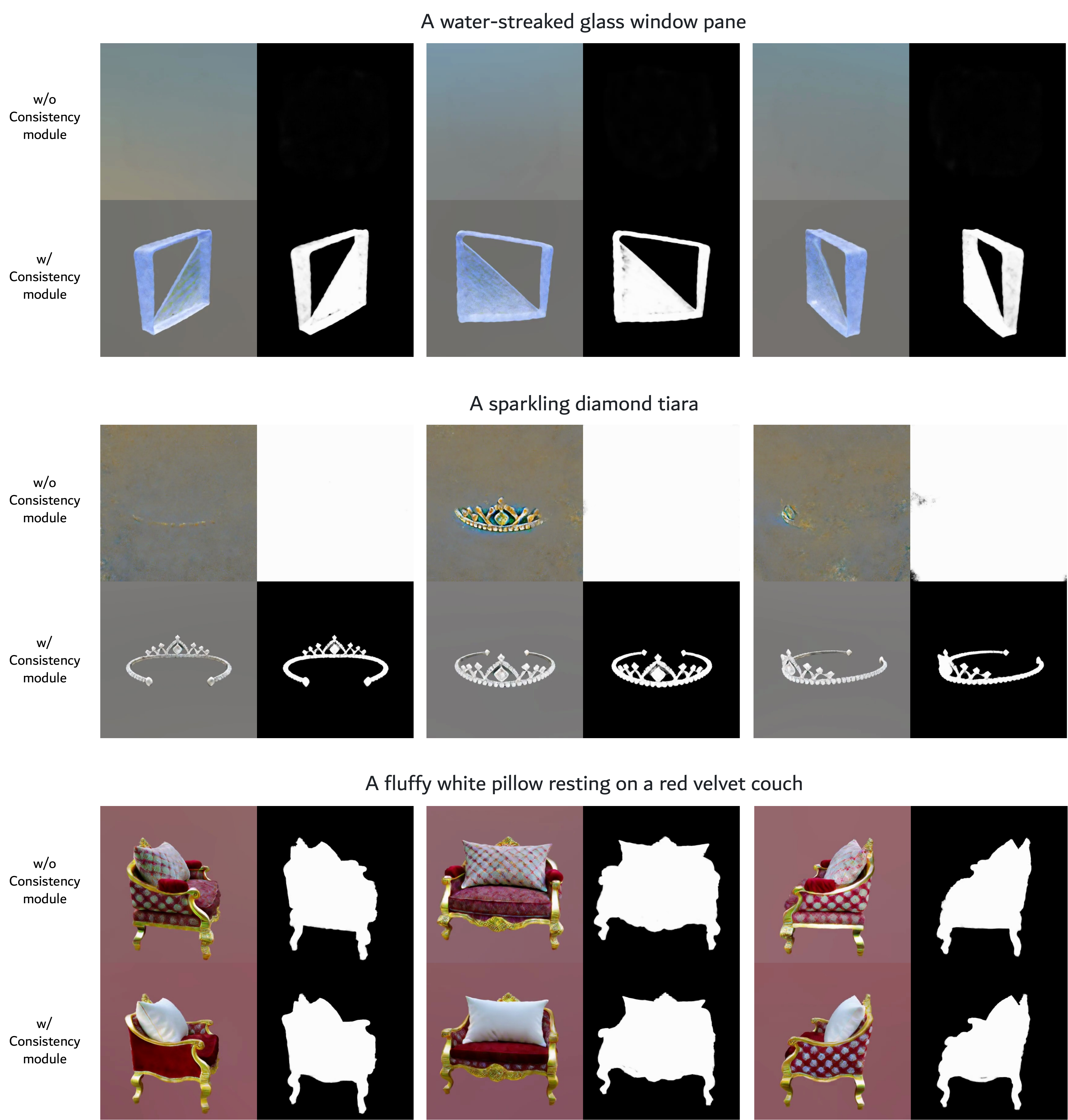}
\caption{Visual comparisons of ablation study. Various cases show that our image-conditioned module exhibits abilities to avoid density vanishing, excessive density generation, and color inconsistency. The left and right parts of the image denote the rendering and the density of the 3D content, respectively.}
\label{fig:abl}
\end{figure}

To show how the image-conditioned module effectively utilizes consistency modeling within a robust, plug-in architecture, we conducted a case study. 
The first row of Figure~\ref{fig:abl} demonstrates how viewpoint inconsistencies in text-conditioned diffusion models can lead to instability when rendering special materials like glass during optimization, often resulting in a loss of density. However, by continuing to constrain cross-view consistency during training, the addition of the image-conditioned module ameliorates such density collapse. Furthermore, when faced with certain prompts (e.g. the second row of Figure~\ref{fig:abl}), the textual inconsistency of diffusion models gradually accumulates erroneous excess density. As these erroneous densities tend not to match the 3D prior in the image-conditioned module, they can be suppressed by our extra module during optimization. Additionally, the third row of Figure~\ref{fig:abl} illustrates the color drift and blending issues that emerge with 2D diffusion guidance. Our module, which facilitates cross-view consistency, effectively addresses these challenges as well.

\section{Conclusion}
\label{sec:conclusion}

In this paper, we propose to optimize a neural radiance field by distilling the score of a text-conditioned multi-view diffusion model and an image-based novel view diffusion model.
Our method is based on the observation of view inconsistency in existing text-based multi-view generative models.
We propose to explicitly enforce a constraint between views by leveraging an image-based diffusion model for supervision.
Thus, unlike prior work, our method can produce accurate densities. 
Currently, our work uses two diffusion models that introduce additional parameters. 
Future work might explore the design of a diffusion model that can generate multiple views given a text as well as generate a novel view given an image.

%%%%%%%%%%%%%%%%%%%%%%%%%%%%%%%%%%%%%%%%%%%%%%%%%%%%%%%%%%%%%%%%%%%%%
%\section*{References} % needed on some systems
\bibliography{mybibfile}

\end{document}